\documentclass[runningheads]{llncs}
\usepackage[T1]{fontenc}
\usepackage{graphicx}
\usepackage{booktabs}
\usepackage[misc]{ifsym}

\usepackage[table]{xcolor}
\usepackage{graphicx}
\usepackage{subcaption}
\usepackage{float}
\usepackage{amsmath}
\usepackage{algorithm, algorithmicx, algpseudocode}
\usepackage{algpseudocode}
\usepackage{tabularx}
\usepackage[version=4]{mhchem}
\usepackage{amssymb}

\algblock{AgentLoop}{EndAgentLoop}
\algblock{Sampling}{EndSampling}
\newcommand{\mathcomment}[1]{\hfill \textit{// #1}}

\usepackage{mwe}
\begin{document}

\title{De Novo Molecular Design Enabled by Direct Preference Optimization and Curriculum Learning}

\titlerunning{De Novo Molecular Design via DPO \& CL}

% If the full title of your paper is short enough to also fit in the running head, you can omit the abbreviated paper title here. You can check as follows: if you comment out the \titlerunning line, something will appear in the header of all odd-numbered pages of your PDF from page 3 onward. This something is either the full title (in which case all is well), or the error message "Title Suppressed Due to Excessive Length". If this error message appears, you're going to want to provide an abbreviated title within the \titlerunning command, because if you won't do it, Springer will do it for you.

%N.B.: Author information (both in the \author{} and \authorrunning{} command) should only be present in the Camera-Ready Version of your paper. The version that you initially submit for review, ought to be double-blind. So, when initially submitting your paper, use:
%\author{Author information scrubbed for double-blind reviewing}

\author{Junyu Hou}

% You may leave out the orcidID information, if you want to.
% Use \corr to indicate the corresponding author. Note the spacing around the \corr command. Only one author can be the corresponding author.

%N.B.: comment out the \authorrunning{} command for the double-blind version of your paper submitted for review. Later, if your paper is accepted, use the command for the Camera-Ready Version.
\authorrunning{De Novo Molecular Design via DPO \& CL}

% If there are two authors, write 'A.L. Benjamin and C.C. Broadus Jr.'
% If there are more than two authors, '[...] et al.' is used.

\institute{Nanjing University 
\email{221220151@nju.edu.cn}}

% \institute{Fictional Southern University, Savannah GA 31404, USA \email{\{a.l.benjamin,a.a.patton\}@fsu.fake}
% \and
% Fictional West Coast University, Long Beach CA 90840, USA \email{ccb@fwcu.fake}
% \and
% Secondary European Affiliation, Tiergartenstr. 17, 69121 Heidelberg, Germany

\maketitle              % typeset the header of the contribution

\begin{abstract}
De novo molecular design has extensive applications in drug discovery and materials science. The vast chemical space renders direct molecular searches computationally prohibitive, while traditional experimental screening is both time- and labor-intensive. Efficient molecular generation and screening methods are therefore essential for accelerating drug discovery and reducing costs. Although reinforcement learning (RL) has been applied to optimize molecular properties via reward mechanisms, its practical utility is limited by issues in training efficiency, convergence, and stability.
To address these challenges, we adopt Direct Preference Optimization (DPO) from NLP, which uses molecular score-based sample pairs to maximize the likelihood difference between high- and low-quality molecules, effectively guiding the model toward better compounds. Moreover, integrating curriculum learning further boosts training efficiency and accelerates convergence. A systematic evaluation of the proposed method on the GuacaMol Benchmark yielded excellent scores. For instance, the method achieved a score of 0.883 on the Perindopril MPO task, representing a 6\% improvement over competing models. And subsequent target protein binding experiments confirmed its practical efficacy. These results demonstrate the strong potential of DPO for molecular design tasks and highlight its effectiveness as a robust and efficient solution for data-driven drug discovery.
%The code is available at \url{https://github.com/programer6497/MolDPO}.

\end{abstract}

\section{Introduction}

De novo molecular design is one of the core tasks in fields such as catalyst design, energy materials design, and pharmaceutical research, aiming to generate novel molecules from scratch that satisfy specified physicochemical properties and biological activity requirements \cite{mandal2009rational}. This process plays a pivotal role in new drug discovery, materials science, and synthetic chemistry. Traditional methods for candidate molecule screening and optimization typically rely on extensive experimental synthesis and biological assays, which are both time- and labor-intensive and require substantial financial investment \cite{dimasi2016innovation}. Moreover, the chemical space is astronomically vast—estimated to contain more than \(10^{60}\) potential molecules \cite{bohacek1996art} —making exhaustive exploration by manual means virtually impossible. Consequently, computer-aided drug design (CADD) has emerged as a prominent research focus, leveraging mathematical models, statistical techniques, and advanced computational technologies to efficiently search and optimize within this enormous chemical space \cite{schneider2005computer,gomez2018automatic,stokes2020deep,zhavoronkov2019deep}.

In recent years, molecular conditional generation has played a critical role in drug development and materials design \cite{meyers2021novo}, and reinforcement learning (RL) has gradually been introduced into the field of molecular design \cite{olivecrona2017molecular,popova2018deep,jin2020multi}. By employing molecular scoring functions as reward signals, RL enables generative models to continuously adjust and improve their generation strategies toward predefined objectives—such as enhancing biological activity, optimizing physicochemical properties, and improving synthetic accessibility—thereby demonstrating enormous potential \cite{simm2020reinforcement,zhou2019optimization}.

However, RL-based methods still face several challenges:  
\textbf{(1) Convergence Challenges and Training Instability}: The high-dimensional and non-convex nature of molecular generation makes RL models prone to slow convergence and local optima. For instance, REINVENT \cite{blaschke2020reinvent} exhibits volatile policy updates and noisy rewards, requiring extensive training before reliably generating molecules that satisfy multiple objectives.  

\textbf{(2) Exploration Inefficiency and Limited Coverage}: The vast chemical space limits the effectiveness of traditional RL approaches, which often get trapped in narrow structural regions. DrugEx \cite{liu2021drugex}, for example, generates molecules meeting specific activity criteria but lacks sufficient scaffold diversity.  

\textbf{(3) Multi-Objective Optimization and Reward Design Challenges}: Designing effective reward functions for molecular optimization is complex and often requires empirical tuning. For example, to simultaneously maximize logP, TPSA, and structural similarity, some studies have employed multi-layered, nonlinear composite reward models, increasing both implementation complexity and limiting generalizability across different tasks \cite{jin2018junction,segler2018generating,sanchez2018inverse}.

To address these challenges, we draw inspiration from two established methodologies in machine learning. Direct Preference Optimization (DPO), originally developed in NLP, has shown strong optimization capabilities in reinforcement learning tasks by leveraging paired samples to optimize likelihood differences, eliminating the need for explicit reward modeling \cite{rafailov2024direct,cheng2024decomposed}. Meanwhile, Curriculum Learning, which gradually increases task complexity, has been adopted to enhance molecular generation \cite{bengio2009curriculum,hacohen2019power}. By starting with simpler tasks and progressively optimizing bioactivity, physicochemical properties, and synthetic feasibility, this approach improves model learning \cite{narvekar2020curriculum,guo2022improving}.We contend that integrating DPO with curriculum learning can both accelerate model convergence and substantially improve the overall performance of the generated molecules.

\begin{figure}[!ht]
    \centering
    \includegraphics[width=1.0\textwidth]{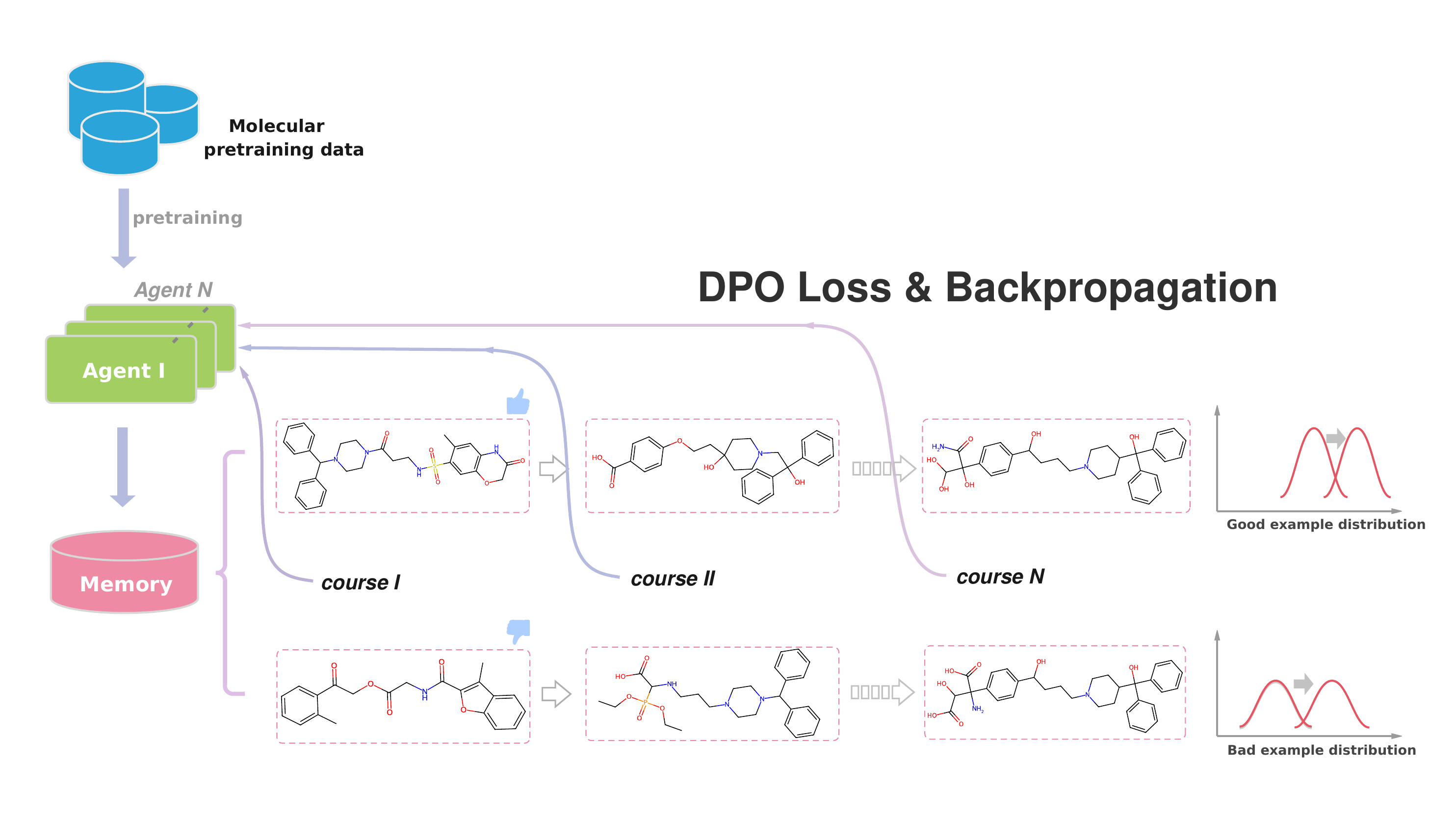}
    \caption{\textbf{Structure of the DPO+Curriculum Learning model.}
    The model is initially pre-trained, followed by optimization using Direct Preference Optimization. As curriculum learning progresses, the molecular scores of the collected compounds steadily increase while the distinction between superior and inferior molecules gradually narrows. Ultimately, the process yields molecules that meet the predefined quality criteria.}
    \label{moudle}
\end{figure}

In this study, we first employ traditional autoregressive training to train a prior model, which is then assigned to four agent models. These agent models are responsible for sampling and constructing paired samples, which are subsequently trained using DPO and curriculum learning to optimize the molecular generation process, ultimately yielding molecules that satisfy the desired properties. We evaluate our method on the Guacamol benchmark \cite{brown2019guacamol} and target protein binding experiments, where experimental results demonstrate that our approach achieves superior performance across multiple evaluation metrics, validating its effectiveness in molecular conditional generation tasks.

The primary objective of this study is to investigate the application of DPO combined with curriculum learning in molecular conditional generation, aiming to improve the molecular generation process through a more efficient and stable approach. By integrating these two methods, we aim to improve molecular discovery and optimization efficiency while providing strong technological support for drug development. Our main contributions are threefold:

\begin{itemize}
    \item We propose a novel de novo molecular design framework that combines Direct Preference Optimization (DPO) with curriculum learning.

    \item Our method has achieved high scores on the GuacaMol benchmark and demonstrated outstanding performance in target protein docking experiments.

    \item The proposed framework exhibits strong potential for scalability in terms of multi-objective optimization, training stability, and computational efficiency.   
\end{itemize}

\section{Related Work}
In recent years, computationally driven de novo molecular design has witnessed rapid advancements, primarily evolving along three key directions: \textbf{(1) continuous optimization of reinforcement learning frameworks}, \textbf{(2) cross-domain adaptation of large language models (LLMs)}, and \textbf{(3) efficiency improvements in preference learning paradigms}. These innovations have enhanced chemical space exploration and multi-objective optimization, laying a stronger foundation for drug discovery.

\subsection{Reinforcement Learning-Based Molecular Generation}
Reinforcement learning (RL)-based generative models have established a systematic paradigm for molecular design\cite{liu2023drugex,zhou2019optimization}. Among them, REINVENT\cite{loeffler2024reinvent} integrates recurrent neural networks (RNNs) with policy gradient algorithms to enable targeted chemical space exploration. Another notable approach\cite{olivecrona2017molecular} leverages prior knowledge to constrain the reward function, optimizing molecular properties while maintaining synthetic feasibility. However, traditional RL approaches face challenges when handling high-dimensional chemical spaces, including inefficient policy updates and susceptibility to local optima.

\subsection{Curriculum Learning Strategies}
To enhance training efficiency in complex tasks, researchers have introduced curriculum learning frameworks into de novo molecular design. Guo et al. \cite{guo2022improving} proposed a strategy that gradually increases task difficulty: the model initially focuses on generating simpler chemical structures, thereby establishing a solid foundation, and then progressively tackles more challenging optimization tasks. This staged approach not only accelerates convergence but also improves the diversity and quality of the generated molecules, demonstrating the effectiveness of curriculum learning in refining generative models for molecular design.

\subsection{Large Language Models for Molecular Generation}
Large Language Models (LLMs) have recently been applied to molecular design, offering novel strategies for molecule generation. Liu et al. \cite{liu2023chatgpt} explored the adaptation of ChatGPT for molecular tasks, demonstrating its ability to capture chemical patterns and generate valid molecular representations through language modeling. Similarly, Hu et al. \cite{hu2023novo} introduced MolRL-MGPT, which integrates a GPT-based generative strategy with reinforcement learning to enhance molecular diversity and optimize target-directed properties. These studies highlight the promising potential of LLMs to provide scalable and effective approaches for molecular design.

\subsection{Preference Optimization in Molecular Design}
Preference learning techniques provide an efficient pathway for strategy optimization in molecular generation. Rafailov et al. proposed direct preference optimization (DPO)\cite{rafailov2024direct}, which employs implicit reward modeling to bypass the complexity of explicit reward function design in traditional RL. This paradigm has recently been successfully adapted to molecular design\cite{gu2025aligning,cheng2024decomposed}: Widatalla et al. utilized experimental data to construct preference pairs, enabling DPO to directly optimize protein stability\cite{widatalla2024aligning}. Experimental results show that ProteinDPO performs exceptionally well in protein stability prediction and demonstrates strong generalization capabilities for large proteins and multi-chain complexes. This suggests that it has effectively learned transferable insights from its biophysical alignment data.

\section{Methodology}
Our molecular generation framework is built upon three core technical components integrated through a structured training pipeline:
(1) Pretraining establishes chemical validity by learning SMILES syntax from large-scale datasets;
(2) Direct Preference Optimization (DPO) replaces reward modeling with contrastive learning to align generation with target objectives;
(3) Curriculum Learning introduces progressive difficulty levels for gradual chemical space exploration.
To synergistically combine these components, we design a two-stage training procedure: pretraining initializes molecular priors, followed by DPO fine-tuning guided by curriculum-constructed preference pairs. The following subsections detail each component.

\subsection{Pretrain on large molecular dataset}
In this study, we adopt the same model architecture as MolRL-MGPT by building a multi-agent GPT model with 8 layers and 8 attention heads \cite{hu2023novo}. Two distinct prior models were pre-trained on different datasets: one on the GuacaMol dataset (a subset of ChEMBL) for benchmark evaluation and another on the ZINC dataset (containing approximately 100 M molecules) for general-purpose molecular generation tasks \cite{irwin2005zinc}. The primary objective during pretraining is to enable the model to deeply learn the syntax rules of SMILES representations, thereby allowing it to generate valid SMILES structures one character at a time and effectively capture the distribution of the chemical space. This pretraining strategy not only demonstrates the model’s capability in efficiently generating chemically valid molecules but also lays a solid foundation for subsequent task-specific optimization and performance enhancement.

To rigorously train our prior models on SMILES representations, we employ an autoregressive framework that decomposes each sequence into a series of incremental prediction tasks. Let \( S = (c_1, c_2, \dots, c_L) \) denote a SMILES sequence of length \(L\), where each \(c_i\) represents a character from the SMILES vocabulary \( \mathcal{V} \). In our autoregressive training approach, we generate training pairs \((x_i, y_i)\) for \(i = 1, 2, \dots, L-1\), where
\begin{equation}
    x_i = (c_1, c_2, \dots, c_i) \quad \text{and} \quad y_i = (c_1, c_2, \dots, c_i, c_{i+1}).
\end{equation}
The model parameterized by \(\theta\) learns a conditional probability distribution \(P_\theta(c_{i+1} \mid c_1, \dots, c_i)\) such that the joint probability of the sequence can be expressed as:
\begin{equation}
P_\theta(S) = \prod_{i=1}^{L} P_\theta(c_i \mid c_1, \dots, c_{i-1}),
\end{equation}
with the convention that \(P_\theta(c_1 \mid \cdot) = P_\theta(c_1)\).

The training objective is to minimize the cross-entropy loss over the entire training set, which for a single sequence is given by:
\begin{equation}
\mathcal{L}(\theta) = - \sum_{i=1}^{L-1} \log P_\theta(c_{i+1} \mid c_1, \dots, c_i).
\end{equation}
This objective encourages the model to assign high probabilities to the correct next character at each step. Parameter updates are performed using gradient descent:
\begin{equation}
\theta \leftarrow \theta - \eta \nabla_\theta \mathcal{L}(\theta),
\end{equation}
where \(\eta\) is the learning rate. This autoregressive framework enables the model to learn the syntax rules of SMILES representations, thereby generating valid molecular structures character by character.

The model trained on the Guacamol dataset for approximately 3 hours, achieving an 97\% validity rate for the generated molecular structures, demonstrating that it had effectively learned the SMILES generation rules. The model trained on the ZINC dataset for approximately 70 hours, achieving a 99.6\% validity rate for the generated molecules, further validating its generalization capability. All pretraining experiments were conducted on a single A100 GPU.

\subsection{DPO for Molecular Optimization}

Building upon the pretrained molecular generation capability, we introduce Direct Preference Optimization (DPO) to align molecular generation with chemical preferences. DPO is a contrastive learning approach that optimizes the generation policy without explicitly modeling a reward function. Instead of using reinforcement learning with human feedback (RLHF) methods that first train a reward model and then optimize the policy using algorithms like PPO, DPO directly optimizes the policy by enforcing preference constraints.

The training data consists of triplets \((x, y_w, y_l)\), where:  
\( x \) represents the input.  
\( y_w \) is the preferred (or "winning") response.  
\( y_l \) is the less preferred (or "losing") response.  

DPO is built upon the idea that an optimal policy \(\pi^*\) should satisfy the following preference ratio constraint:

\[
\frac{\pi^*(y_w | x)}{\pi^*(y_l | x)} = \exp(r(y_w | x) - r(y_l | x))
\]

where \( r(y \mid x) \) is an implicit reward function that ranks different responses. Instead of explicitly learning this reward function, DPO directly optimizes the policy ratio by defining the following log preference probability:

\[
\log \sigma(\beta \cdot (\log \pi_\theta(y_w | x) - \log \pi_\theta(y_l | x)))
\]

where \( \sigma(z) = \frac{1}{1 + e^{-z}} \) is the sigmoid function, and \( \beta \) is a temperature hyperparameter controlling sensitivity to preference differences.

The final DPO objective function is:

\[
\mathcal{L}(\theta) = \mathbb{E}_{(x, y_w, y_l) \sim D} \left[ \log \sigma(\beta \cdot (\log \pi_\theta(y_w | x) - \log \pi_\theta(y_l | x))) \right]
\]

In our molecular generation task, there is no explicit input $x$. Furthermore, whereas other DPO tasks often require human annotation of preference samples, which is costly, our task does not rely on human annotations to determine response quality. Instead, we leverage a chemical computation library to evaluate the quality of generated molecules, thereby streamlining the preference learning process. 

%This objective ensures that the model assigns higher probabilities to preferred responses while penalizing less preferred ones, without requiring explicit reward modeling or reinforcement learning techniques. By using DPO, we effectively guide our molecular generation model to produce outputs that align more closely with domain-specific chemical preferences.

\subsection{Curriculum Learning for Structured Molecular Optimization}
To address the challenge of learning complex chemical spaces, we integrate DPO with curriculum learning. Curriculum learning is a machine learning strategy inspired by the human learning process, where the model begins with simple tasks and gradually progresses to more complex ones. By organizing training samples in order of increasing difficulty, the model builds a solid foundation with easier examples, leading to more efficient learning, better generalization, and ultimately enhanced performance on challenging tasks.

Aligned with this progressive approach, our pair construction process incrementally increases the difficulty of the learning task. Initially, the score gap between the superior and inferior samples is large, making it straightforward for the model to distinguish high-quality molecules. As training advances, this gap is gradually reduced, requiring the model to discern more subtle differences. This strategy reinforces the curriculum learning paradigm and refines the model's fine-grained discrimination of molecular quality, ultimately enhancing the validity of the generated compounds.

Furthermore, we adopt a multi-stage learning mechanism to enhance model performance. Specifically, during training, high-quality molecules collected by the model are stored in memory. Subsequently, all agents are reinitialized to the pre-trained model and continue training by constructing new sample pairs from the high-scoring molecules in memory. The primary objective of this strategy is to mitigate potential biases introduced during early exploration, preventing the model from converging to suboptimal solutions. By leveraging previously identified high-quality molecules, the model can effectively restart its learning process in a more optimized direction, ultimately improving the quality of generated molecules.

\begin{figure}[!ht]
    \centering
    \includegraphics[width=0.8\textwidth]{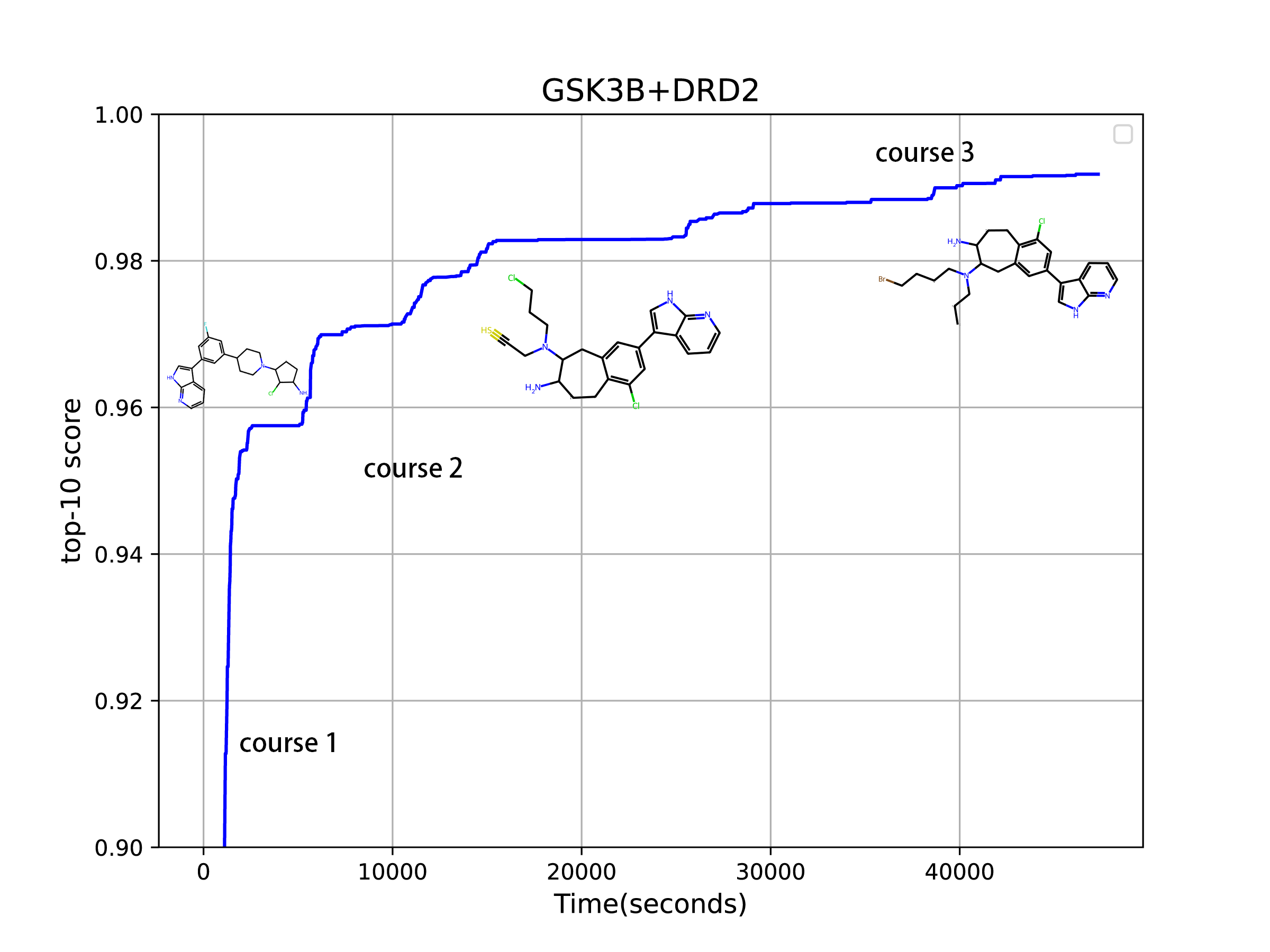}
    \caption{\textbf{In the GSK3B+DRD2 docking experiment, the model achieved good performance through curriculum learning.} In Course 1, the model learns the fundamental requirements of the task. In Course 2, it fine-tunes the molecular scaffold. In Course 3, it adjusts functional groups to optimize molecular structures.}
    \label{courses}
\end{figure}

As illustrated in the figure \ref{courses}, our training process is divided into three stages. In the first stage, the model learns the fundamental requirements of the task and rapidly identifies high-scoring molecules from the vast chemical space. In the second stage, the model fine-tunes molecular scaffolds to further refine its understanding. Finally, in the third stage, the model modifies functional groups based on the optimal molecular scaffolds stored in memory, further optimizing molecular structures. This multi-stage approach significantly enhances molecular design efficiency, leading to a remarkable score of 0.993 on the GSK3B+DRD2 task.

\subsection{Training Procedure}
Integrating DPO with curriculum learning, we design a two-stage training protocol, as illustrated in Figure \ref{moudle}. The process begins with pre-training to obtain the prior model, followed by reinforcement fine-tuning using DPO and curriculum learning.

In the pre-training phase, the model learns to generate valid SMILES strings and capture the chemical space distribution, forming a prior model.

During reinforcement fine-tuning, the agents—initialized from the prior—generates molecules, which are evaluated by a task-specific scoring function. Preference pairs are then constructed by selecting the top-k highest-scoring molecules (k varies across agents) as “preferred samples” and randomly sampling lower-quality ones as “dispreferred samples”. These pairs are used to optimize the policy via DPO, progressively aligning the model's generation strategy with the target objectives.

As training progresses, the scores of the generated molecules steadily improve while the gap between preferred and dispreferred samples narrows, indicating an increase in training difficulty. Initially, the model makes broad scaffold-level adjustments to identify promising frameworks; as scaffolds stabilize, it shifts to fine-tuning functional groups to further optimize molecular properties.

The DPO training process code is as follows:

\begin{algorithm}[!ht]
\caption{Direct Preference Optimization (DPO) Training}
\begin{algorithmic}[1]
\Procedure{DPO\_Train}{$F_{\text{score}}, k$}
    \Statex \textbf{Initialize:}
    \State Load pretrained prior $p_{\text{ref}} \gets \text{GPT}(\theta_{\text{prior}})$
    \State Initialize agents $\{ \pi_i \}_{i=1}^N$ with $\theta_i \sim \mathcal{N}(0, 0.02)$
    
    \For{$t \gets 1$ to $T$}
        \AgentLoop
        \For{each agent $\pi_i$}
            \Sampling
            \State $\mathcal{D}_i \gets \text{SampleSMILES}(\pi_i, m_{\text{batch}})$
            \State $\mathbf{s} \gets F_{\text{score}}(\mathcal{D}_i)$
            \State $\mathcal{M} \gets \text{UpdateMemory}(\mathcal{D}_i, \mathbf{s})$
            \EndSampling
            
            \State \textsc{Positive Selection}: 
            \[
            \mathbf{x}^w \sim p_{\mathcal{M}}(x) \propto \exp(s/\tau) 
            \mathcomment{top-weighted historical samples}
            \]
            \State \textsc{Negative Selection}:
            \[
            \mathbf{x}^l \sim \text{Uniform}(\mathcal{D}_t) \mathcomment{current batch negatives}
            \]
            \State Compute log-ratios for each sample:
            \[
            \log r_{\theta}(x) = \log\pi_{\theta}(x) - \log\pi_{\text{ref}}(x)
            \]
            \State Optimize loss: 
            \[
            \mathcal{L}_{\text{DPO}} = -\mathbb{E}\left[ \log \sigma\left( \beta (r_{\theta}(x^w) - r_{\theta}(x^l)) \right) \right]
            \]
            \State Gradient step: $\theta \gets \theta - \eta \nabla_{\theta}\mathcal{L}_{\text{DPO}}$

        \EndFor
        \EndAgentLoop
        
        \State Log $\max\mathcal{M}.\mathbf{s}$, $\text{top}_k\text{-mean}(\mathcal{M}.\mathbf{s})$
    \EndFor
    
    \State \Return $\text{Top}_k(\mathcal{M}.\mathbf{x}, \mathcal{M}.\mathbf{s})$
\EndProcedure
\end{algorithmic}
\end{algorithm}

%Through this two-stage training process, our model is able to further improve the generation quality based on reinforcement learning, and better align with the task objectives through Direct Preference Optimization.

\section{ Experiments}
\label{headings}

To validate the effectiveness of our model, we designed and conducted a series of experiments, including the Guacamol benchmark evaluation, target protein binding experiments, and impact analysis. The experimental results demonstrate that our model is not only capable of handling classical molecular design tasks but also performs exceptionally well in tasks that are more closely aligned with real-world drug discovery. Furthermore, the impact analysis, which examines model performance under different parameter settings, helping us identify the optimal parameter settings.

\subsection{GuacaMol benchmark}

\subsubsection{Guacamol Introduction}

Guacamol Benchmark, proposed by BenevolentAI in 2019 \cite{brown2019guacamol}, is a standardized framework for evaluating molecular generation models in terms of diversity, synthetic feasibility, and goal-directed optimization. It comprises 20 tasks covering key challenges in molecular design.

These tasks can be broadly categorized into rediscovery and similarity-based optimization, isomer generation, and molecular property balancing. Additionally, multi-parameter optimization (MPO) tasks focus on improving physicochemical properties of known drugs, while SMARTS-constrained tasks enforce structural constraints. Lastly, scaffold hopping and decorator hopping tasks assess the model’s ability to modify core structures and substituents.

\subsubsection{Baselines}
To comprehensively evaluate our approach, we compare it against several representative baselines:
\textbf{SMILES LSTM} \cite{segler2018generating}: An LSTM-based model trained via maximum likelihood estimation to generate SMILES strings.
\textbf{Graph GA} \cite{jensen2019graph}: A graph-based genetic algorithm that optimizes molecular structures through crossover and mutation.
\textbf{Reinvent} \cite{blaschke2020reinvent}: A model combining recurrent neural networks with reinforcement learning, using reward functions to enhance both bioactivity and physicochemical properties.
\textbf{GEGL} \cite{ahn2020guiding}: An approach that integrates graph neural networks with reinforcement learning to directly optimize molecular graphs.
\textbf{MolRL-MGPT} \cite{hu2023novo}: A hybrid model that fuses GPT-based generative strategies with reinforcement learning to boost molecular diversity and target-specific performance.

\subsubsection{Experimental Details}

First of all, we pre-trained the model using the training set provided by Guacamol. The pre-training was conducted for 15 epochs on a single A100 GPU over a duration of 3 hours. After pre-training, the model achieved a molecular validity of \textbf{97\%}, demonstrating its high accuracy in molecular structure generation.

Subsequently, we further trained the model on 20 tasks from the Guacamol benchmark to evaluate its performance across different objectives. The \textbf{hyperparameter} settings used in this stage were as follows:  \textbf{Batch size} = 50, \textbf{n\_steps} = 1000, \textbf{num\_agents} = 4, \textbf{Learning rate} = 1e-4, \textbf{Memory size} = 1000.

% Additionally, we introduced several special settings for specific tasks to enhance optimization performance:
% \begin{itemize}
%     \item In the Sitagliptin MPO task, we increased the learning rate to 5e-4 to avoid being trapped in suboptimal solutions.
%     \item In the \ce{C9H10N2O2PF2Cl} task, to discover all 250 isomers, we expanded the memory size to 2000 and assigned four agents to learn molecules ranked in the top 100, 250, 500, and 1000, ensuring broader search coverage.
% \end{itemize}

The complete training of the 20 benchmark tasks required 60 hours, whereas MolRL-MGPT took 400 hours under the same conditions (both on a single A100 GPU). Our model demonstrated a training speed nearly 6 times faster, highlighting the advantages of DPO’s stable training and faster convergence, which significantly reduces training costs.

\definecolor{lightblue}{RGB}{230, 240, 250}

\begin{table}[!ht]
\centering
\caption{ Scores of DPO and baselines on the GuacaMol benchmark. (All task scores are rounded to three decimal places.)}
\begin{tabularx}{1.0\textwidth}{l|XXXXX>{\columncolor{lightblue}}X}
\toprule
\textbf{ } & SMILES-LSTM & GraphGA & Reinvent & GEGL & MolRL-MGPT & \textbf{DPO\&CL} \\  % Table header
\midrule
Celecoxib rediscovery  & \textbf{1.000}  & \textbf{1.000}  & \textbf{1.000}  & \textbf{1.000}  & \textbf{1.000}  & \textbf{1.000}   \\ 
Troglitazone rediscovery   & \textbf{1.000}  & \textbf{1.000}  & \textbf{1.000}  & 0.552  & \textbf{1.000}  & \textbf{1.000}   \\ 
Thiothixene rediscovery & \textbf{1.000}  & \textbf{1.000}  & \textbf{1.000}  & \textbf{1.000}  & \textbf{1.000}  &  \textbf{1.000}  \\ 
Aripiprazole similarity & \textbf{1.000}  & \textbf{1.000}  & \textbf{1.000}  & \textbf{1.000}  & \textbf{1.000}  &  \textbf{1.000}  \\ 
Albuterol similarity  & \textbf{1.000}  & \textbf{1.000}  & \textbf{1.000}  & \textbf{1.000}  & \textbf{1.000}  & \textbf{1.000}   \\ 
Mestranol similarity   & \textbf{1.000}  & \textbf{1.000}  & \textbf{1.000}  & \textbf{1.000}  & \textbf{1.000}  &  \textbf{1.000}  \\ 
\ce{C11H24}  & 0.993  & 0.971  &  0.999  & \textbf{1.000}  &  \textbf{1.000}  &  \textbf{1.000}  \\ 
\ce{C9H10N2O2PF2Cl}   &  0.879  & 0.982  &  0.877 & \textbf{1.000}   & 0.939 &  \textbf{1.000}  \\ 
Median molecules 1 & 0.438  & 0.406  & 0.434  & \textbf{0.455}  & 0.449  &  \textbf{0.455}  \\ 
Median molecules 2  & 0.422  &  0.432  &  0.395  & \textbf{0.437}  & 0.422  &  0.422  \\ 
Osimertinib MPO  & 0.907  & 0.953  & 0.889  & \textbf{1.000}  & 0.977  & 0.990   \\ 
Fexofenadine MPO  & 0.959 & 0.998 & \textbf{1.000} & \textbf{1.000} & \textbf{1.000} & \textbf{1.000} \\ 
Ranolazine MPO  & 0.855 & 0.920 & 0.895 & 0.933 & 0.939 & \textbf{0.950} \\ 
Perindopril MPO & 0.808 & 0.792 & 0.764 & 0.833 & 0.810 & \textbf{0.883}  \\ 
Amlodipine MPO  & 0.894 & 0.894 & 0.888 & 0.905 & \textbf{0.906} & \textbf{0.906} \\ 
Sitagliptin MPO & 0.545 & \textbf{0.891} & 0.539 & 0.749 & 0.823 & 0.838 \\ 
Zaleplon MPO & 0.669 & 0.754 & 0.590 & 0.763 & 0.790 & \textbf{0.797} \\
Valsartan SMARTS & 0.978 & 0.990 & 0.095 & \textbf{1.000} & 0.997 &  0.994 \\ 
deco hop& 0.996 & \textbf{1.000} & 0.994 & \textbf{1.000} & \textbf{1.000} & \textbf{1.000} \\ 
scaffold hop  & 0.998 & \textbf{1.000} & 0.990 & \textbf{1.000} & \textbf{1.000} & \textbf{1.000}  \\ \hline
Total & 17.340 & 17.983 & 16.350 & 17.627 & 18.052 & \textbf{18.235} \\ \bottomrule % Total row
\end{tabularx}
\label{guacamol}
\end{table}

As shown in the table \ref{guacamol}, our model achieved the best performance on multiple tasks, with its overall score surpassing other baselines. Specifically, our method outperformed existing approaches on 16 out of 20 benchmark tasks, demonstrating a clear advantage in molecular generation. Compared to GEGL, our model exhibited higher stability across diverse tasks, achieving consistently superior performance rather than excelling in only a subset of cases.

Our model’s effectiveness is particularly evident in challenging tasks. For instance, in Perindopril MPO, it significantly outperformed existing methods by a margin of 0.05, highlighting its robustness in complex molecular design. Another compelling example is the Ranolazine MPO task, where MolRL-MGPT, the previous state-of-the-art model, improved upon its closest competitor by only 0.006, suggesting that performance on this task had reached a plateau. However, our approach further improved upon MolRL-MGPT by an additional 0.011, demonstrating that our model can break through existing performance bottlenecks and further optimize molecular generation outcomes.

These results indicate that our model possesses a strong learning capability, effectively handling molecular generation tasks and producing high-quality molecules that meet target requirements. Furthermore, these findings validate the effectiveness of the DPO method in optimizing molecular generation, providing a solid foundation for future research. 

\subsection{Molecular Generation for High Binding Affinity to Target Proteins}

In this experiment, we utilized a prior model pretrained on the ZINC dataset. The evaluation was conducted on six tasks: JNK3, GSK3B, DRD2, and their pairwise combinations (JNK3+GSK3B, JNK3+DRD2, GSK3B+DRD2).

The model performance was evaluated using the oracle function provided by TDC. For multi-objective optimization tasks, we used the arithmetic mean of the individual target scores as the final score to assess the overall performance of the generated molecules across multiple targets.

\begin{table}[!ht]
\caption{The scores of the generated molecules on JNK3, GSK3$\beta$, DRD2, and pairwise combination tasks.}
\centering
\begin{tabular}{|c|c|c|c|}
\hline
\textbf{ } & \textbf{top 1} & \textbf{top 10 mean} & \textbf{top 100 mean}\\ \hline  % Table header
\textbf{JNK3} & 1.000 & 1.000 & 1.000  \\ \hline
\textbf{GSk3B} & 1.000 & 1.000 & 1.000 \\ \hline
\textbf{DRD2} & 1.000 & 1.000 & 1.000 \\ \hline
\textbf{JNK3+GSK3B} & 0.944 & 0.943 & 0.938 \\ \hline
\textbf{JNK3+DRD2} & 0.925 & 0.925 & 0.920 \\ \hline
\textbf{GSK3B+DRD2} & 0.993 & 0.992 & 0.989 \\ \hline
\end{tabular}
\label{docking}
\end{table}

As shown in the table \ref{docking}, our model successfully generated molecules with strong binding potential to JNK3, GSK3B, and DRD2, demonstrating its effectiveness in molecular generation tasks. These results indicate that our model not only performs well on the Guacamol benchmark but also excels in real-world drug discovery tasks, providing a solid foundation for future research in molecular design and generation.

\subsection{Impact Analysis}

This study suggests that multiple factors, including the learning rate, the number of agents, the sampling-to-training ratio, and the DPO parameter $\beta$, may influence model performance. Through preliminary analysis, we identified that the number of agents and the sampling-to-training ratio have a particularly significant impact. To validate this hypothesis, we conducted a systematic impact analysis experiment focusing on these two key parameters. The experimental results demonstrate that appropriately adjusting the number of agents and the sampling-to-training ratio can significantly enhance model performance, providing valuable theoretical insights and practical guidance for further model optimization.

\subsubsection{Agents Num}

\begin{figure}[!ht]
    \centering
    \begin{subfigure}{0.48\textwidth}
        \centering
        \includegraphics[width=\textwidth]{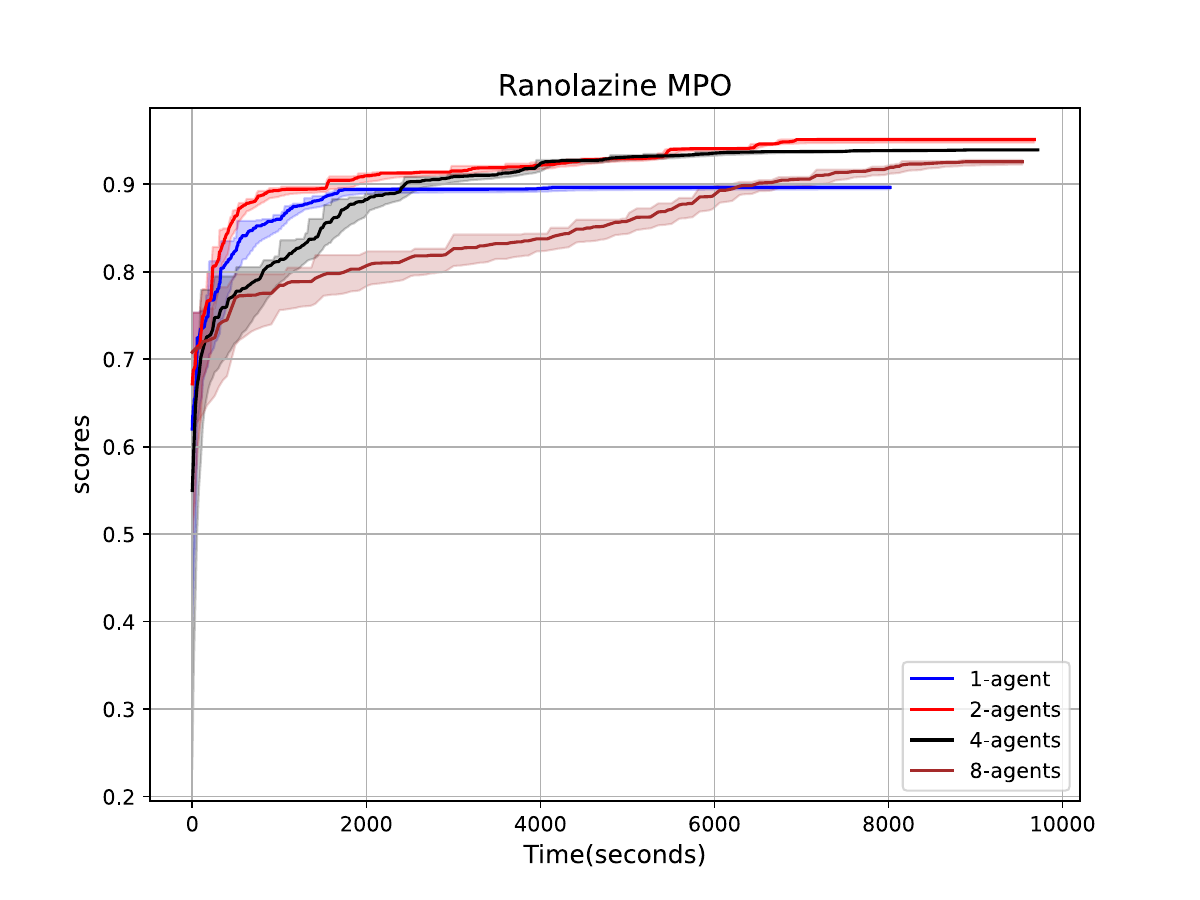}
    \end{subfigure}
    \hfill
    \begin{subfigure}{0.48\textwidth}
        \centering
        \includegraphics[width=\textwidth]{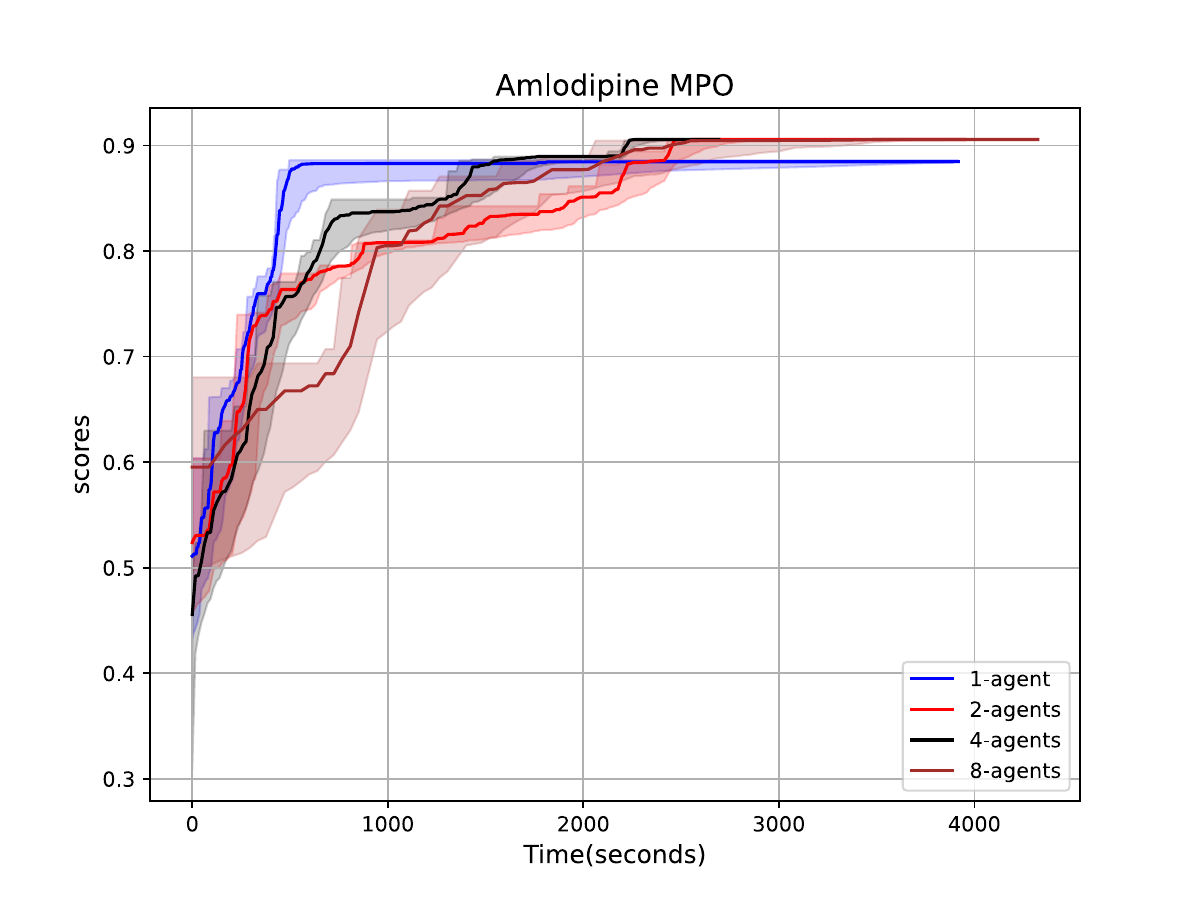}
    \end{subfigure}
    \caption{Model performance on Ranolazine MPO and Amlodipine MPO tasks under different numbers of agents. (The curve represents the Top-10 score, while the shaded region indicates the score distribution of the top 100 molecules.)}
    \label{agents}
\end{figure}

Experimental results in figure \ref{agents} indicate that the model achieves optimal performance when employing 2 to 4 agents. When the number of agents is too small, the model's expressive capacity is limited, making it difficult to effectively learn complex patterns. Conversely, an excessive number of agents may introduce redundant information and increase optimization complexity, thereby lowering the model’s performance upper bound and slowing down convergence. Moreover, the number of agents also affects the score distribution of molecules stored in memory: fewer agents lead to a more concentrated distribution, whereas a larger number of agents result in a more dispersed distribution. Due to differences in the learning configurations of individual agents, their capabilities diverge as training progresses, leading to an increasingly diverse molecular distribution. This broader distribution facilitates the optimization of DPO training and enhances the model’s generalization capability.

\subsubsection{Sampling-to-Training Ratio}

\begin{figure}[!ht]
    \centering
    \begin{subfigure}{0.48\textwidth}
        \centering        \includegraphics[width=\textwidth]{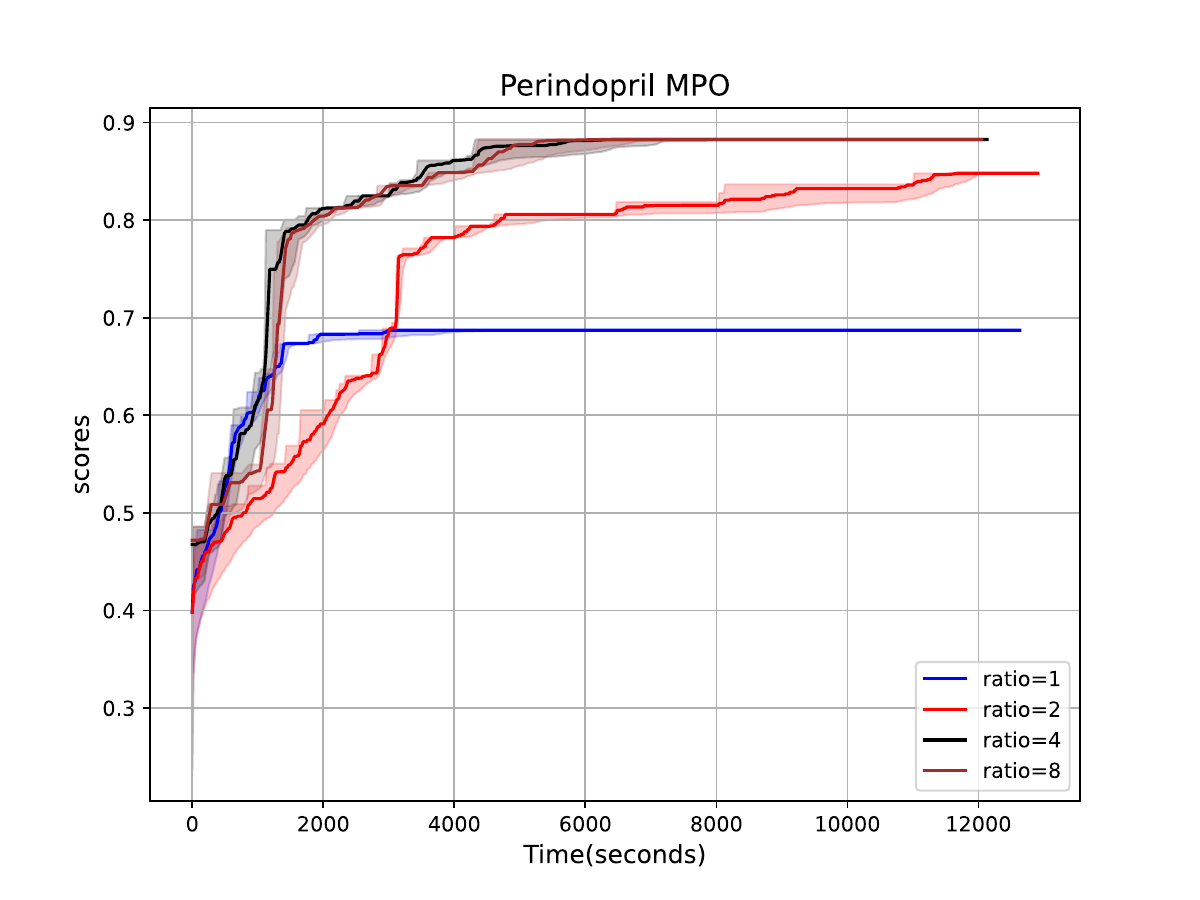}
    \end{subfigure}
    \hfill
    \begin{subfigure}{0.48\textwidth}
        \centering
    \includegraphics[width=\textwidth]{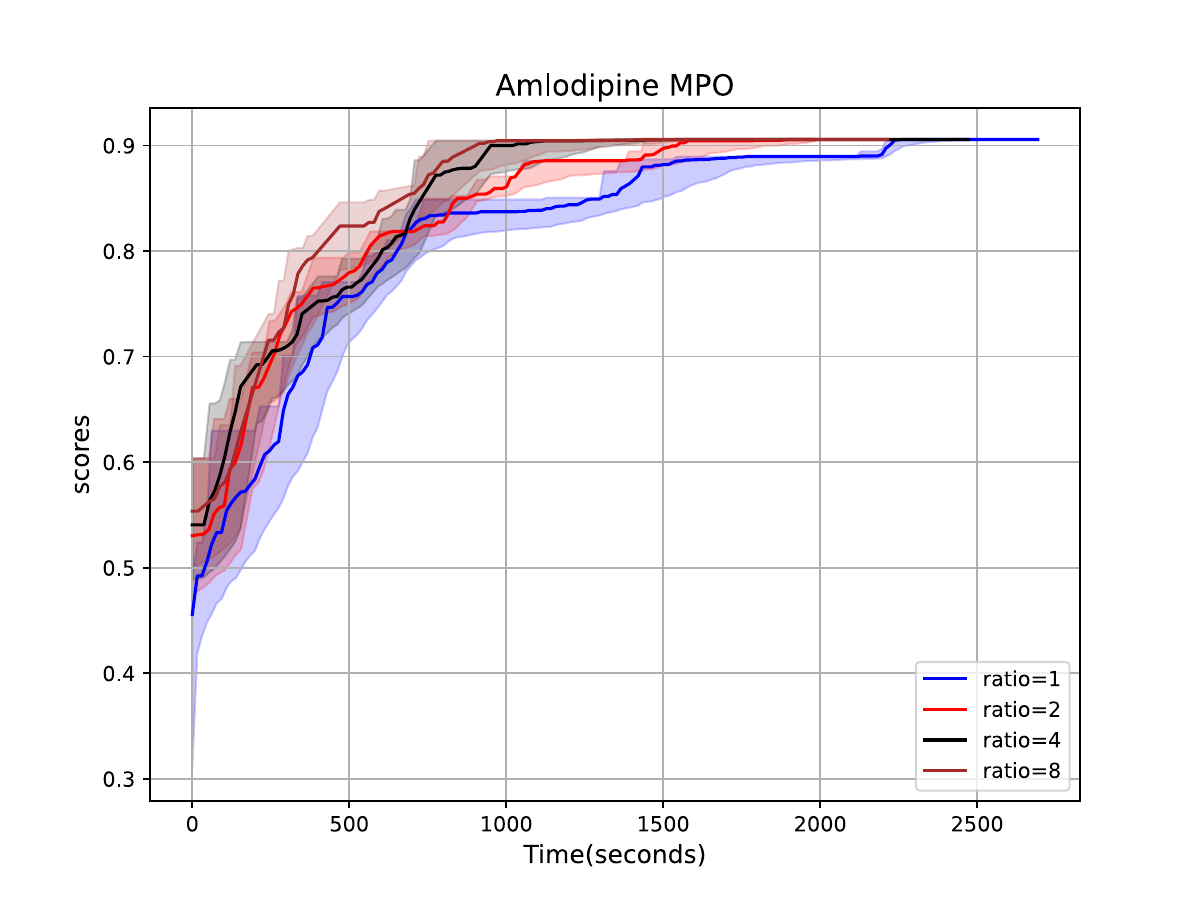}
    \end{subfigure}
    \caption{Model performance on Perindopril MPO and Amlodipine MPO tasks under different Sampling-to-Training Ratios. (The curve represents the Top-10 score, while the shaded region indicates the score distribution of the top 100 molecules.)}
    \label{ratio}
\end{figure}

As shown in the figure \ref{ratio}, increasing the sampling-to-training ratio within a certain range can improve the model’s performance upper bound and accelerate convergence. If the ratio is too small, the weights assigned to a few high-quality molecules obtained by chance become excessively large, causing the model to shift towards them—even when they do not represent the optimal direction—ultimately leading to a suboptimal solution. Conversely, if the ratio is too large, the model remains in the sampling phase for an extended period, collecting a large number of redundant molecules while lacking sufficient training. This results in inadequate gradient updates and significantly prolongs the model’s convergence time.

\section{Conclusion}
This study proposes a molecule generation method based on DPO and curriculum learning, and achieves favorable experimental results on the Guacamol benchmark and several target tasks. The experiments demonstrate that the proposed method has significant advantages in tasks that generate molecules with specified properties.

\begin{credits}
% \subsubsection{\ackname} A bold run-in heading in small font size at the end of the paper is
% used for general acknowledgments, for example: This study was funded

\subsubsection{\discintname}
The authors have no competing interests to declare that are relevant to the content of this article.

\end{credits}
%% Note that this preceding line implies that you store your BibTeX references in a file called 'mybibliography.bib'. If you instead store your references in a file with a different name, for instance 'references.bib', the preceding line should read '\bibliography{references}'. Whatever you do, DO NOT put the file name extension .bib inside the \bibliography command; this will trip up LaTeX compilers. 
%

\bibliography{main}

\end{document}